# Black-Box Segmentation of Electronic Medical Records

Hongyi Yuan[1,2], Sheng Yu[1,2]
[1]Center for Statistical Science, Tsinghua University, Haidian, Beijing, China; [2]Department of Industrial Engineering, Tsinghua University, Haidian, Beijing, China

**Abstract**

*Electronic medical records (EMRs) contain the majority of patients' healthcare details. It is an abundant resource for developing an automatic healthcare system. Most of the natural language processing (NLP) studies on EMR processing, such as concept extraction, are adversely affected by the inaccurate segmentation of EMR sections. At the same time, not enough attention has been given to the accurate sectioning of EMRs. The information that may occur in section structures is unvalued. This work focuses on the segmentation of EMRs and proposes a black-box segmentation method using a simple sentence embedding model and neural network, along with a proper training method. To achieve universal adaptivity, we train our model on the dataset with different section headings formats. We compare several advanced deep learning-based NLP methods, and our method achieves the best segmentation accuracies (above 98%) on various test data with proper training corpus.*

**Introduction**

Natural language processing (NLP)[1] technologies are essential for extracting information from electronic medical records (EMRs), where the majority of healthcare details are written in free text. Most studies of NLP techniques and the development of software for EMR processing have been focusing on concept extraction[2], assertion analysis[3], relation extraction[4], and phenotyping[5] at various granularity, while not sufficient attention has been given to the automatic sectioning of EMRs. The section structure provides essential context for information extraction from EMRs. For example, a disorder identified from the family history section should not be tagged as the patient's current condition, and a medication identified from the allergy section is unlikely one that the patient currently takes. This kind of background information from the sections should be handled carefully to avoid incorrect phenotyping. It also requires explicitly knowing the section structure of the EMR notes because the sections' background information is typically not repeated in their sentences.

Explicitly programming for the section structures can be difficult. Many early EMR systems, such as the one demonstrated in the Medical Information Mart for Intensive Care[6] (MIMIC-III) data, treated the whole note as a single document and did not have dedicated database fields, file tags, or even a uniform format for the sections. It is possible to identify patterns and develop sophisticated rules based on combinations of upper-case words, empty lines, and cue words to section the notes, but the work is time-consuming and prone to error if not tested sufficiently. For multi-institutional studies, writing rules for the systems of each participating institution can be prohibitive. Modern EMR systems typically have standard templates or formats for the notes, making the sections easy to identify using tags or rules. However, coding for multiple institutions still requires significant efforts. Additionally, regulations to protect patient privacy may not allow the programmer to examine the notes and develop rules for every institution. Therefore, machine learning-based sectioning algorithms that can adapt to all kinds of formats, so that they can be used as black boxes, are of great value.

Machine learning-based EMR sectioning can be modeled as a sequential sentence classification problem, with sections being the classes. Two textual characteristics of EMRs should be noted to achieve accurate sectioning. The first characteristic is that sentences of different sections exhibit different semantic patterns (e.g., word distribution). Naïve Bayes[7] and logistic classifiers[8] are classic machine learning tools to utilize the word distributions, but they have been surpassed by latest deep learning[9] techniques, such as word embeddings[10,11] and pre-trained language models[12,13]. The other characteristic to enhance the classification accuracy is the underlying pattern of section transitions. Doctors are trained to write the EMRs in a relatively standard logic and order. In other words, the transition probability of the sections exhibits a strong pattern. This pattern can be effectively used by classic methods, such as the hidden Markov model[14] and conditional random fields[15] (CRF). For deep learning, the bidirectional long short-term memory[16] (BiLSTM) layer and the CRF layer can be used for sequential classification along with the embedding layer and achieve higher accuracy than traditional machine learning[15].

In this paper, we propose and compare several deep learning-based models for the automatic sectioning of EMRs. We use the MIMIC-III data as test material. Hand-crafted rules are developed to segment the discharge notes to create the gold-standard labels for training and testing. To achieve universal adaptivity, we generate multiple versions of the notes by modifying the section headings' format to train the models, and we also test their performances on the various formats. We show that our method achieves nearly 100% accuracy on EMRs segmentation with a training corpus of mixed formats of the section headings.

**Related Work**

In this work, the EMR segmentation problem is modeled as sequential sentence classification. Segmentation is realized based on accurate sort of sentences, not paragraphs. In recent years, prior works focused on the scientific sequential sentence classification, especially on biomedical sequential sentences.

Prior work presented a public dataset for sequential sentence classification called PubMed 200k RCT[17]. The dataset contains short PubMed article abstract texts with sentences labeled by one of the following classes: background, objective, method, result, and conclusion. There were some naive techniques such as logistic regression and simple neural networks tested on the dataset. Several related works tested their methods on this dataset.

Existing work proposed a hierarchical neural network method. Word embeddings were fed into a BiLSTM and attention mechanism network to get sentence embeddings. Recurrent Neural Network (RNN) and CRF layers were implemented to classify the sentence embeddings[18]. This method had good performance on the short PubMed abstract text, while when dealing with long texts such as EMRs, the stacked network would be too large to train. Instead of using BiLSTM and attention mechanism to present sentences, Convolutional Neural Network (CNN) was used to acquire sentence representations[19]. Methods such as Sent2Vec[20] and fastText[21] were also used to encode sentences which were used as inputs to different classification neural network structures[22]. An approach named SR-RCNN was developed in an early work[23]. The proposed method used RNN and TextCNN[24] to encode sentences, and a multilayer RNN and a CRF layer were used to predict the label for each sentence. This work found that applying attention mechanism to sentence embeddings would not improve the classification accuracy. The benefit of adding a CRF layer to help predict label sequences depends on the specific dataset.

Using fine-tuned Bidirectional Encoder Representations from Transformers (BERT) to classify sequential sentences was also explored[25]. In nowadays NLP field, pre-trained language models had drawn much attention due to their surprising performance. Proper fine-tuned BERT models achieved state-of-art results on different NLP tasks[12,13]. This work showed that the fine-tuned BERT model with a simple linear layer at the output has a state-of-art result tested on the PubMed 200k RCT dataset. It was also illustrated that a CRF layer is redundant because the attention mechanism inside BERT has already captured the context information.

There was no work focusing on the segmentation of EMRs or the sequential sentence classification on long texts to the best of our knowledge. The existing methods degrade when handling long texts. Our work deals with sequential sentence classification on long EMR texts instead of short PubMed abstract texts for the first time.

**Methods**

This section introduces the proposed method for EMR segmentation, and the classification model mainly consists of two parts: the sentence embedding part and the sequential sentence classification part. We use an unsupervised sentence embedding method, and we also compare it with the simple word average method and advanced pre-trained BERT method. The sequential sentence classification methods use a neural network to predict an appropriate label for each sentence. Sequential sentence classification methods also consider the inter-dependence between sequential sentences which implies the orders of EMRs' writing pattern. To show the superiority of our method, we test several comparison methods. How the comparison methods adapt to the EMR segmentation task is also presented. The details of the models are as follows in this section.

**Sentence Embedding Part** We firstly approach this problem by embed sentences in EMRs into a high-dimension vector space. To appropriately represent sentences and use a small number of computational resources, the Smooth Inverse Frequency[26] (SIF) is applied to construct sentence embeddings. This sentence embedding method is unsupervised and straightforward based on word embeddings, word existing probability, and Singular Value Decomposition / Principal Component Analysis (SVD / PCA). For simplicity, the theoretical derivation is omitted. Intuitively, the ideas behind this method are that the inverse of word existing frequency can alleviate the influence of common words such as stop words. PCA / SVD is used on the matrix of concatenated sentence embeddings in one note to find the joint direction of embedding vectors, the first principal component (the first singular vector). By subtracting this, the mutual information is eliminated, and it is believed to help distinguish different sentence vectors.

This method is verified to outperform the RNN or LSTM based methods such as skip-thoughts[27] on some downstream tasks such as text similarity or text entailment.

In the following narratives, the word embeddings are denoted as $e_{ij}$, and the word tokens are denoted as $w_{ij}$, and sentence embeddings are denoted as $s_j$. The subscript $i$ and $j$ stand for the $i$th word in the $j$th sentence in one EMR. The word embeddings $e_{ij} \in \mathbb{R}^d$ are unsupervised trained on the whole EMR corpus using methods word2vec[10] or GloVe[11]. The probability of a word existing in the entire corpus is also calculated as:

$$p(w) = \frac{n_w}{n_{all}},$$

where $n_w$ is the times of word $w$ existing in the whole corpus, and $n_{all}$ is the number of all words. Then, the embedding of a sentence is calculated as:

$$s_j = \frac{1}{n} \sum_{i=1}^{n} \frac{\alpha}{\alpha + p(w_{ij})},$$
$$s_j = s_j - u^T u\, s_j,$$

where $s_j \in \mathbb{R}^d$ and $u$ is the first singular vector of the matrix $[s_1, s_2, \cdots s_j, \cdots]$ containing all the sentence vector embeddings in one EMR note. $u$ is the result of performing SVD on the sentence vector embedding matrix. $\alpha$ is a small preset parameter for smoothing. The final sentence embedding will serve as inputs for the sequential sentence classification network.

**Sequential Sentence Classification Part** After acquiring sentence embeddings, a neural network is implemented to predict sentence sequences tags. The neural network is composed of RNN layers. The output of the network is the softmax probability distributions of labels for each sentence.

Although EMR notes may have different section structures, we assume that the EMR notes from various resources share the same writing order pattern. As for the sentences in each section, those who share the same label always exist successively and seldom exist after/before sentence sequences of some specific labels. For example, in MIMIC-III discharge summary notes, sentences of Social History are always presented before those of Family History. Therefore, a CRF layer is added to capture this kind of pattern of the writing order.

CRF is widely used in Named Entity Recognition (NER) tasks. CRF predicts the transition scores from one label to another, and the objective of the CRF layer is to find the label sequence that maximizes the sum of transition scores and emission scores (which is the probabilities of predicted labels). The CRF layer is designed for capturing the writing orders of sentence sequences in EMRs.

**Comparison Method** To compare with the proposed model above, the trivial one is to use word embedding average to present sentence vectors. An alternative method is to use a pre-trained BERT model to encode sentences. BERT-based methods do well in a variety of NLP tasks. A sentence with the [CLS] token concatenated at the beginning is fed into the pre-trained BERT model. Then the output vector of [CLS] token is used as the sentence embedding. In this way, the sentence embedding is fixed, so the BERT model will not be fine-tuned while training.

We also compare another approach of using pre-trained BERT to achieve sequential sentence classification. Inspire by a prior work[25], sentence sequences are used as BERT inputs rather than sole sentences. In the input sentence sequence, a [SEP] token is concatenated at the end of each sentence, and a [CLS] token is also inserted at the beginning of the whole input sequence. We use the BERT output vectors of [SEP] tokens to represent the corresponding sentences. Then the sentence embeddings are fed into a neural network to realize classification. Three different network structures are compared: linear networks, RNN with a CRF layer, and attention mechanism with a CRF layer. The attention mechanism is defined as follows. Consider a sequence of BERT output vectors for words in a sentence $\{v_{w_1}, v_{w_2}, \cdots, v_{w_i}, \cdots\}$ and the corresponding [SEP] token vector $v_{sep}$,

$$\alpha_i = \frac{exp\big((W_q v_{sep})^T W_k v_{w_i}\big)}{\sum_{j=1}^{N} exp\big((W_q v_{sep})^T W_k v_{w_j}\big)},$$
$$h_i = \sum_{j=1}^{N} \alpha_i\, W_v v_{w_j}.$$

Where matrix $W_q$, $W_k$, $W_v \in \mathbb{R}^{d \times d}$ are trainable parameters and $h_i$'s are the attention vector representations for sentences. Under this method, the pre-trained BERT model will be fine-tuned along with the neural network during training.

The BERT model is made of multiple transformer layer, which is mainly constructed by attention mechanism. In the next sentence distinguishing task that pre-trains the BERT model, the [SEP] token is inserted between two input sentences to learn both sentences' structures and relations. So intuitively, after fine-tuning the attention weight in the downstream sequential sentence classification task, the [SEP] token embedding is believed to be capable of capturing the information and the transition mode in sentence sequences.

**Experiments**

Our objective is to segment EMRs, which is to assign sentences to preset labels. This section introduces the EMR corpus used in the experiment and how to form our training and testing corpus. The rules to construct different formats of training and testing samples will be presented in detail. Other methods are tested on the synthetic dataset, and results will be compared and analyzed.

**MIMIC-III Discharge Summary Notes** MIMIC-III database contains various patient information from 2001 to 2012 in the Intensive Care Unit (ICU) in the Beth Israel Deaconess Medical Center in Boston, Massachusetts. In the MIMIC-III dataset, the discharge summary notes are free texts and comprise the patients' necessary information, conditions, medical histories, exams, treatments, etc. There are more than 50,000 notes of about 40,000 patients in MIMIC-III (Some patients were admitted to ICUs more than once). We use the discharge summary notes in the MIMIC-III dataset to generate our training and testing data.

**Data Preparation** In the origin discharge summary notes, although they are free texts, there are headings for a variety of sections. Then we use these original section headings to assign sentences to different preset labels using rule-based methods. With doctors' help, one discharge summary note is mainly segmented into the following 25 parts of contents (Table 1). Some contents may not exist in different discharge summary notes. These 25 parts of contents form the labels for sequential sentence classification in the training and testing dataset.

**Table 1.** 25 Parts of Discharge Summary Notes

| Admission Date | Discharge Date | Date of Birth |
|---|---|---|
| Sex | Service | Allergies |
| Attending | Chief Complaint | Major Surgical or Invasive Procedure |
| History of Present Illness | Review of System | Past Medical History |
| Social History | Family History | Physical Exam |
| Pertinent Result | Hospital Course | Medication on Admission |
| Discharge Medications | Discharge Disposition | Facility |
| Discharge Diagnosis | Discharge Condition | Discharge Instruction |
| Follow-up Instruction | | |

The sentences in discharge summary notes are assigned to the preset labels above according to two rules:

***Rule 1.*** *If the original section heading is a substring of one specific label, all sentences in the origin section are tagged with the same label and vice versa;*

***Rule 2.*** *If an original section heading (e.g., LAB) cannot match any label by Rule 1, sentences in this origin section will be tagged with the label of the section ahead (e.g., sentences in LAB always exists after those sections are tagged with Physical Exam, then LAB is regarded as a part of Physical Exam).*

By these two rules, all the sentences in EMRs are assigned with the corresponding label. We manually check several generated samples, and sentences are correctly tagged.

After tagging sentences, to clean up the sentences and narrow down the token vocabulary, first of all, words in texts are converted to their lowercase. Ignoring the information of the letter case is rational in our setting. Secondly, in MIMIC-III discharge summary notes, to protect the privacy of patients, dates, names, locations, phone numbers, and all other private information are masked. The mask token contains the category of the information and a unique code, or a mask is an irrational substitute (e.g., [**2118-6-7**] as a mask for a date). These kinds of masks will cause trouble in the embedding part because, in the training phase, they will enlarge the token vocabulary, and they are hard to acquire useful vector embedding. To alleviate this problem, we substitute these masks with a universal mask for each information category (e.g., date masks are substituted by [date]).

Besides, patients in ICU are carefully monitored. Thus, various exams or medications are given, and different numbers and units in EMRs record the exam results and medication dosages. Many of these numbers only exist once in the entire corpus and will also make our token vocabulary too large. Those numbers that only exist in testing samples will also cause the similar problem as the original masks. When classifying sentences, the specific number or unit is redundant. It is logical to replace all the numbers or units in the corpus with the same mask. Furthermore, all the symbols (e.g., +) in sentences are removed.

It should be noted that the pre-trained BERT model limits the maximum input sequence to 512. Unfortunately, sentences tagged by Physical Exam or Medication are extremely long, exceeding 512 words. It is due to these sentences are tabular contents which are many concatenated word-number-unit triples in discharge summary notes. These kinds of sentences are cut into sub-sentences within 512 words to adapt to the input limit of the pre-trained BERT model.

Following the above steps, one discharge summary note generates one sample with a cleaned sentence sequence and corresponding labels. Samples are split into training samples and testing samples to validate our model.

**Experiment Settings** Word embeddings are trained via word2vec on the entire corpus. The dimension of the word vector is set to 300. The window size is set to 16. The sentence embedding dimension for SIF and averaging method is 300, while for BERT-based models is 768, which is preset for pre-trained BERT models. As for the SIF method, the smooth parameter α is set to 0.001 because the best choice of α is between 0.0001 to 0.001 according to the paper[26]. BiLSTM is applied to the RNN layer of classification network parts. Using the pre-trained BERT model, we select the BioBERT model from HuggingFace's Transformer[28]. The BioBERT shares the same parameter size and structure as the original BERT, and it is especially pre-trained on the medical corpus from PubMed. When training different models, batch sizes are adjusted due to the limited memory of computation resources.

To test the generalization of models and achieve universal adaptivity, we generate four types of samples from each note by modifying their section headings. The four types of samples are:

***Type 1.*** *Removing the original section headings;*

***Type 2.*** *Maintaining the original section headings;*

***Type 3.*** *Substituting the original headings with their corresponding preset labels;*

***Type 4.*** *Adding corresponding labels before and after each section to form an XML format sample.*

We try three kinds of training corpus to discover how different formats of section heading of training sample impact the trained models' ability. In the first kind of training corpus, we only generate Type 1 samples. In the second kind, we only generate Type 2 samples. In the third kind, we randomly generate the above four types of training samples, and these four types are evenly distributed in this mixed kind of training corpus. When testing the generalization of trained models, we respectively test on the above four types of samples.

Three kinds of training corpus are generated from 40,000 randomly selected notes, and testing corpus generated from others.

**Results**

We test six different models, including our proposed one. Through different models, we aim to show that good sentence embedding is critical for our problem and compare our proposed methods with fine-tuned BERT-based methods, which achieve state-of-art performance on the PubMed 200k RCT dataset. All the models are respectively trained on three kinds of training corpus, and their accuracy results on four different types of testing samples are shown as follows (Table 2, Table 3, Table 4).

There are modifications when implementing BERT-based models. When use pre-trained BERT [CLS] tokens to embed sentences, there are too many sentences to embed, and the calculation is time-consuming even without fine-tuning it. Limited to computational resource, BERT [CLS] embeddings with BiLSTM and CRF model only train with 7000 samples and test with 3000 samples.

When using the [SEP] token of BERT output, the inputs are sequential sentences on word granularity, and the word length of one EMR note will surely exceed 512 tokens. Early work[25] illustrated this problem when handling long sequences. To adapt the method to EMRs, we cut one note into several sequential sentence parts whose lengths are within 512 tokens of words. Then the classification model has performed on these sub-sequences rather than the entire sentence sequence of words.

**Table 2.** Model Accuracies Using Training Corpus with Original Section Headings

| Methods \ Sample Types | Type 1 | Type 2 | Type 3 | Type 4 |
|---|---|---|---|---|
| SIF+BiLSTM+CRF** | 0.1419 | **0.9968** | **0.9812** | **0.9461** |
| AVE+BiLSTM+CRF | 0.3296 | 0.8324 | 0.784 | 0.8516 |
| BERT[CLS]+BiLSTM+CRF | 0.2286 | 0.2121 | 0.2141 | 0.1874 |
| BERT[SEP]+Linear | 0.5929 | 0.9522 | 0.9521 | 0.8486 |
| BERT[SEP]+BiLSTM+CRF | 0.5478 | 0.9838 | 0.9836 | 0.9129 |
| BERT[SEP]+ATTN+CRF | **0.6218** | 0.9282 | 0.9243 | 0.8878 |

**Table 3.** Model Accuracies Using Training Corpus without Section Headings

| Methods \ Sample Types | Type 1 | Type 2 | Type 3 | Type 4 |
|---|---|---|---|---|
| SIF+BiLSTM+CRF** | **0.9782** | **0.8229** | **0.8495** | 0.3150 |
| AVE+BiLSTM+CRF | 0.9229 | 0.7687 | 0.7693 | 0.4136 |
| BERT[CLS]+BiLSTM+CRF | 0.5086 | 0.4184 | 0.4148 | 0.3439 |
| BERT[SEP]+Linear | 0.962 | 0.7734 | 0.7834 | 0.6590 |
| BERT[SEP]+BiLSTM+CRF | 0.9731 | 0.6827 | 0.7164 | 0.5574 |
| BERT[SEP]+ATTN+CRF | 0.8425 | 0.7341 | 0.7208 | **0.6723** |

**Table 4.** Model Accuracies Using Training Corpus of Mixed Four Formats Section Headings

| Methods \ Sample Types | Type 1 | Type 2 | Type 3 | Type 4 |
|---|---|---|---|---|
| SIF+BiLSTM+CRF** | **0.9823** | **0.9947** | **0.9953** | **0.9950** |
| AVE+BiLSTM+CRF | 0.8215 | 0.9074 | 0.9158 | 0.9365 |
| BERT[CLS]+BiLSTM+CRF | 0.4482 | 0.4904 | 0.4888 | 0.4307 |
| BERT[SEP]+Linear | 0.9600 | 0.9754 | 0.9753 | 0.9746 |
| BERT[SEP]+BiLSTM+CRF | 0.9701 | 0.9838 | 0.984 | 0.9884 |
| BERT[SEP]+ATTN+CRF | 0.9238 | 0.9659 | 0.9658 | 0.9786 |

*The best results are bold.

** SIF+BiLSTM+CRF is our proposed method, AVE is short for averaging word embeddings, ATTN is short for attention mechanism.

The tables above show that our proposed model gives the best performance. Our model trained on the mixed kind of corpus reaches nearly 100 percent accuracies on different types of testing samples. With the same sequential classification neural network structure, comparing to word-average sentence embeddings, SIF sentence embeddings outperform simple average embeddings by averaging about 10%. SIF sentence embeddings also beat pre-trained BERT embedding by averaging about 40%. These results show SIF embeddings capture more semantic information and help to distinguish sentences from different subsections. The simple average embeddings will suffer from commonly shared words in all the sentences, such as stop words. BERT sentence embeddings without fine-tuning cannot adjust to specific tasks very well and are even worse than the averaged ones.

The models use fine-tuned BERT [SEP] tokens to embed sentences also give good results when training with three kinds of training samples. The models also show good robustness across four types of testing samples. Compared with our proposed model, a slight decrease in performance is observed. It is observed that different neural network structures attached to fine-tuned BERT model have a marginal influence on the performance. The model with the attention mechanism with a CRF layer even slightly degrades in contrast to linear layers.

We find out that models trained on the mixed kind of corpus give the best adaptivity and accuracy through experiments. When models train on a corpus with original headings, they acquire lower accuracy scores on Type 1 samples without original section headings. Models trained on corpus without original headings perform worse on the Type 4 samples of the XML format. This phenomenon shows that section headings contain important information when sorting sequential sentences. Corpus with mixed heading types of training samples helps models capture this information and acquire better generalization ability.

**Discussion**

The quality of sentence embedding is critical for classification. SIF is superior to the simple averaging method. However, it is counter-intuitive that using pre-trained BERT [CLS] embedding vectors results in very low accuracies. These results show that this type of embeddings does not learn the structure and information of a sentence without BERT fine-tuned and cannot precisely represent a sentence. This phenomenon is also illustrated in early work[29].

Fine-tuned BERT has the greatest number of parameters among these six models, and it should have the best performance. However, fine-tuned BERT models do not achieve the highest accuracies on this task. This is partly because we split discharge summary notes to adapt to the input length limit of the BERT model. The context information is lost for those sentences on the edge of each sub-sentence sequence, and the first label in the sub-sequence input is not always the same. Moreover, in some objective sections such as Physical Exam, Past Medical History, and Hospital Course, there are many sentences inside these sections, and these sections may contain words far beyond 512. This will make some BERT input sub-sentence sequences all belong to the same label. These are the bad samples for the CRF layer to learn the sequential labels' pattern because there is no label transition in these samples. These two reasons both make this sequential sentence classification task more difficult for fine-tuned BERT models.

It is also showed that different structures connected to BERT models, RNNs with CRF layer, attention mechanism with CRF layer, and simple linear layer have similar results. The attention mechanism with the CRF layer even performs slightly worse than the linear layer without the CRF layer. The attention inside fine-tuned BERT has already captured the information and dependency between sequential sentences, so the linear layer is sufficient for models' classification ability. Adding more complex structures have the whole model even harder to train.

The formats of training samples have a significant influence on the trained model. Firstly, when training models use samples with original section headings have low accuracy on testing samples without those headings. The low accuracy is due to the trained model categorizing all the successive sentences to the same label. The model relies much on the transition information provided by the original headings. When there is no section heading, models tend to fail. Secondly, when training on the corpus without original subsection headings, it achieves low accuracy on XML format testing samples. In the training stage, models do not learn how to assign the right labels to section headings, especially for those at the end of a section. Therefore, the model will predict a wrong label for the heading after the texts in XML format. This error will mislead the model to give wrong predictions on the successive sentences. Using the mixed kind of training corpus, models have seen different formats of training samples. Various types of section headings as transition signals are captured, resulting in a robust and excellent classification ability.

The proposed method segments EMRs into corresponding parts precisely. We harvest the best result training our proposed model with mixed section headings training corpus through comparison of different models and training methods. It is illustrated the proposed simple model can result in a brilliant outcome. Compared with BERT-based models, our proposed model is easy to train and implemented when the computing resource is limited and there is no need for high-end GPUs.

Our work based on the supervised method relies on a fair amount of labeled data. In the field of healthcare, labeled data is hard to collect, while there are many unlabeled free texts. In future works, we will have further research on sample efficiency or developing an unsupervised or semi-supervised method.

**Conclusion**

In this paper, we present how MIMIC-III discharge summary notes are cleaned and tagged to form our data samples and the idea of approaching the EMR segmentation task as a sequential sentence classification problem. Further, we propose a model and a training method to classify sentences in EMRs and achieve high accuracy and universal adaptability. We compare the existing sentence embedding methods and different deep learning models on this task. When training on a mixed section heading dataset, the proposed model composed of the unsupervised SIF embedding technic and BiLSTM with a CRF layer network achieves the best performance. The model can adapt to various formats of EMRs, and can be used as a black-box method with good generalization ability. It requires fewer computational resources compared with the pre-trained BERT-based models. EMR corpus with a universal structure will help the other medical informatics task such as information extraction and help the doctors access specific EMR subsections conveniently.